\documentclass[letterpaper, 10 pt, conference]{IEEEtran}  %
\usepackage[letterpaper,left=.75in,right=.75in,top=.75in,bottom=.75in]{geometry}

\IEEEoverridecommandlockouts

\usepackage[pdftex]{graphicx}
\graphicspath{{figures/}}
\usepackage{url}
\usepackage{footmisc}
\usepackage{hyperref}
\hypersetup{
	colorlinks=true,
	linkcolor=black,
	citecolor=black,
	filecolor=black,
	urlcolor=black,
}
\usepackage[T1]{fontenc}
\usepackage[utf8]{inputenc}
\usepackage{csquotes}
\usepackage[english]{babel}
\usepackage[export]{adjustbox}
\usepackage{caption}
\usepackage{subcaption}
\usepackage[colorinlistoftodos,backgroundcolor=orange!20,bordercolor=orange]{todonotes}
\usepackage{lipsum}
\usepackage{amsmath}
\usepackage{amssymb}
\usepackage{mathtools}
\usepackage{calc}
\usepackage{pgfplots}
\usepackage{siunitx}
\usepackage{tabularx,booktabs}%
\newcolumntype{C}{>{\centering\arraybackslash}X} %
\usepackage{placeins}%
\usepackage{multirow}

\usepackage{todonotes}

\usepackage[style=ieee,
doi=false,
url=false,
mincitenames=1,
maxcitenames=1,
minbibnames=6,
maxbibnames=6,
backend=biber]{biblatex}  %
\newcommand*\DS{\text}

\title{Exploiting Multi-Layer Grid Maps for Surround-View Semantic Segmentation of Sparse LiDAR Data}
\author{
Frank Bieder$^1$,  Sascha Wirges$^2$, Johannes Janosovits$^1$, Sven Richter$^1$, Zheyuan Wang$^1$\\ and Christoph Stiller$^1$%
\thanks{$^1$Authors are with Institute of Measurement and Control Systems, Karlsruhe Institute of Technology, Karlsruhe, Germany. {}} 
\thanks{$^2$Author is with FZI Research Center for Information Technology, Karlsruhe, Germany.{}}
}

\begin{document}
\maketitle

\IEEEpubid{\begin{minipage}{\textwidth}~\\[12pt] \centering%
	\copyright~2020 IEEE. Personal use of this material is permitted. Permission from IEEE must be obtained for all other uses, including reprinting/republishing this material for advertising or promotional purposes, collecting new collected works for resale or redistribution to servers or lists, or reuse of any copyrighted component of this work in other works.
  \end{minipage}}
  
  \IEEEpubidadjcol
\pagestyle{empty}

\begin{abstract}
In this paper, we consider the transformation of laser range measurements into a top-view grid map representation to approach the task of LiDAR-only semantic segmentation.
Since the recent publication of the SemanticKITTI data set, researchers are now able to study semantic segmentation of urban LiDAR sequences based on a reasonable amount of data. 
While other approaches propose to directly learn on the 3D point clouds, we are exploiting a grid map framework to extract relevant information and represent them by using multi-layer grid maps. 
This representation allows us to use well-studied deep learning architectures from the image domain to predict a dense semantic grid map using only the sparse input data of a single LiDAR scan. 
We compare single-layer and multi-layer approaches and demonstrate the benefit of a multi-layer grid map input. 
Since the grid map representation allows us to predict a dense, 360\si{\degree} semantic environment representation, we further develop a method to combine the semantic information from multiple scans and create dense ground truth grids. 
This method allows us to evaluate and compare the performance of our models not only based on grid cells with a detection, but on the full visible measurement range. 
\end{abstract}
\section{INTRODUCTION}
\label{sec:INTRODUCTION}

\begin{figure}[t!hbp]
\centering
\includegraphics[width =\linewidth]{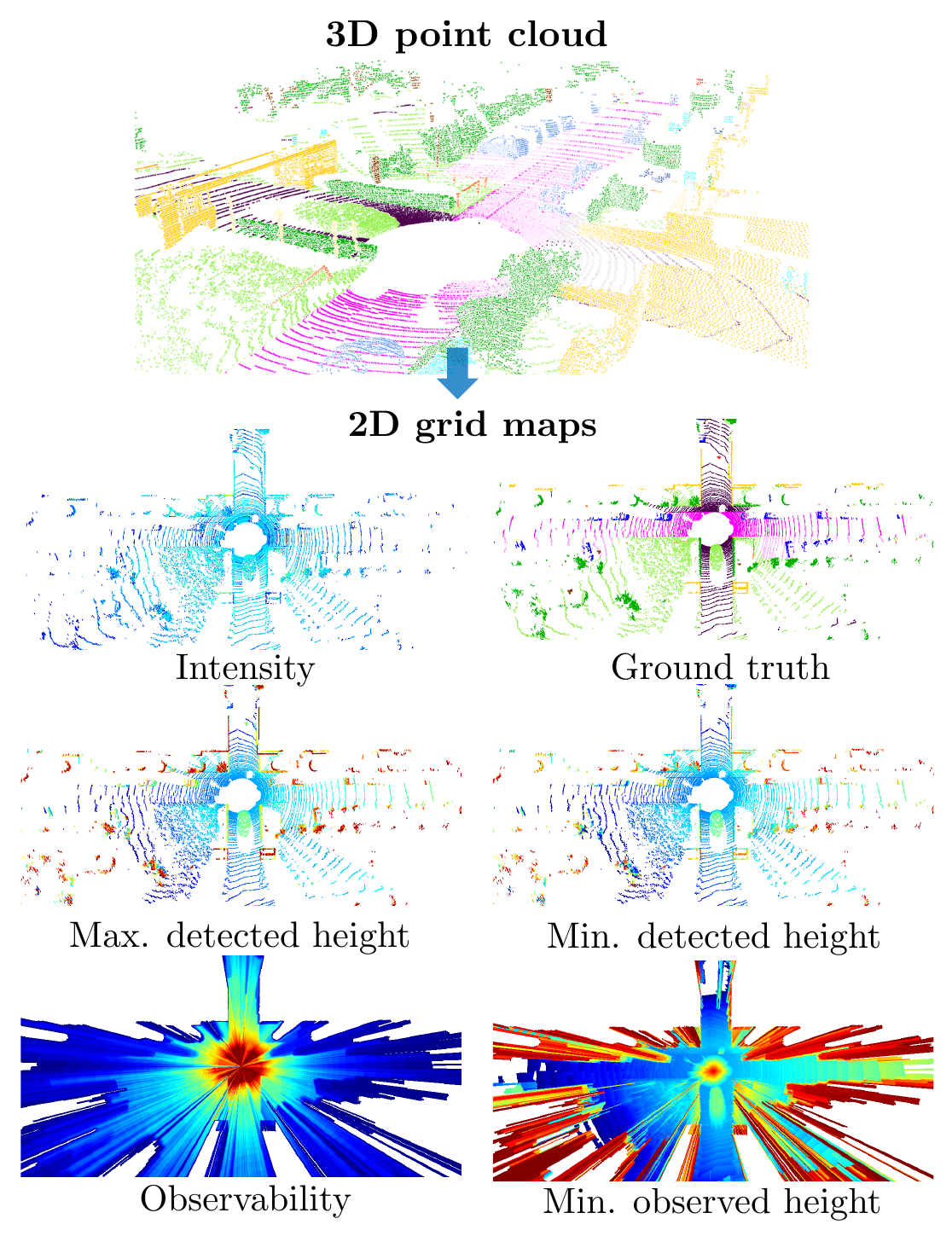}
\caption{
    We transform semantically annotated point sets provided by the SemanticKITTI \cite{Behley2019} data set into a multi-layer grid map representation.
    Using these grid maps we train a pixel-wise semantic labeling with five non-semantic input layers and the semantic grid map as training labels.
}
\label{fig:IntroGridMapLayer}
\end{figure}

A successful deployment of intelligent, mobile systems such as automated vehicles in complex urban scenarios demands an excellent scene understanding and a reliable interpretation of the surroundings.
To reach this goal the car has to be equipped with a range of complementary sensor systems and a sufficient amount of computational resources to extract the relevant information.
Semantic segmentation – assigning a semantic class to each single pixel of an image or point in a point-cloud – is a crucial task in the process of achieving a meaningful virtual representation of a vehicle's environment.
In the last years, deep convolutional neural networks have proven to be very successful in solving multiple computer vision tasks and significant progress has been made in terms of semantic segmentation.
Here, most research has been conducted in the vision domain since the majority of available data sets for semantic segmentation are image-based \cite{BDD100} \cite{Cordts2016}.
Recently, the first large-scale laser-based semantic data set, called SemanticKITTI, was published. This will allow researchers to focus on the semantic classification of LiDAR measurements.
The usage of laser sensors has several advantages: They provide a 360\si{\degree} surround-view of precise depth information, are robust against varying lighting conditions and are able to measure the reflectance of objects. %

\IEEEpubidadjcol
However, the processing and classification of high-resolution 3D point clouds requires high computational effort and most deep learning architectures are optimized to operate on well-structured input representaitions.
Considering this situation, we propose to encode the laser measurements into a multi-layer grid map representation.
In contrast to an unstructured point cloud, the grid map representation is well suited for sensor fusion applications \cite{Nuss2015} \cite{Richter2019}.
This enables us to leverage high-quality deep learning algorithms from the image domain due to the structural properties of the grid maps.
In contrast to the semantic segmentation of LiDAR-based range maps \cite{Milioto2019}, our approach predicts a semantic grid, i.e. a dense environment representation with an equidistant cell structure.

Our approach differs from previous publications by generating the semantic grids with labeled LiDAR scans.
The main contributions of this paper are the following: Firstly, we show that semantic segmentation of grid cells can be learned using sparse input data from LiDAR sensors.
We compare the usage of multi-layer input of different richness and evaluate the results in terms of accuracy and inference time.
Since a main advantage of our approach is the dense prediction of equidistant grid cells, we also propose a method which exploits the semantic information of up to 300 surrounding laser scans to form a dense semantic ground truth and enables up to evaluate the dense predictions of our models properly.

\section{Related Work} \label{sec:related_work}

Conceptually, the presented work is influenced by multiple research fields which we will introduce in the following subsections.

\subsection{Grid Mapping} \label{sec:related_work_grid_mapping}
2D occupancy grid maps were first introduced by Moravec \cite{Moravec1989} and Elfes \cite{Elfes1989}.
Due to the orthographic projection used during mapping, measurements are scale invariant which makes grid maps well-suited for sensor fusion tasks \cite{Nuss2015}.
When grid maps are modeled as regular grids, efficient operations such as unary operations and convolutions can be implemented which enables their use on GPUs \cite{Homm2010} and with machine learning applications (e.g. \cite{Wirges2018Aug}).
Homm et al. \cite{Homm2010} show that range sensor measurements can be mapped into polar grid coordinates and then efficiently transformed to a cartesian grid map.
In addition to occupancy, multi-layer grid maps also encode other information such as average reflected energy, minimum and maximum height of measurements and information on occluded areas obtained via ray-casting.
These different features enable the application of unsupervised methods (e.g. Connected Components Labeling) and supervised learning methods (e.g. convolutional object detectors \cite{Wirges2018Obj}) to grid maps.

\subsection{Benchmarks and Data Sets} \label{sec:related_work_data_sets}
The PASCAL VOC challenges \cite{Everingham2014} were the first benchmarks for the pixel-wise semantic classification of camera images.
Although its pixel-wise annotated images cover many scenarios, the amount of available on-road and traffic images is small. 
Therefore, the KITTI Vision benchmark suite \cite{Geiger2013} was the first data set with benchmarks for automated driving applications.
It was gradually extended with benchmarks for other applications e.g. for visual odometry, scene flow estimation \cite{Menze2015} and object detection.
In 2018, a benchmark for semantic segmentation in camera images \cite{AbuAlhaija2018} was added which provides an alternative to the Cityscapes data set \cite{Cordts2016}.
Recently, Behley et al. \cite{Behley2019} presented a benchmark for the point-wise semantic classification of KITTI range sensor measurements.
The released part of the data set comprises a subset of the KITTI odometry benchmark sequences in which every point in more than 23000 lidar scans was annotated with one of 28 semantic classes. Some of these classes are further divided in \textit{moving} and \textit{non-moving} objects.

\subsection{Semantic Segmentation} \label{sec:related_work_semantic_segmentation}
Semantic segmentation in camera images refers to the task of assigning a semantic label to each image pixel.
Due to availability of large, annotated data sets (e.g. \cite{Everingham2014, AbuAlhaija2018}), Deep Learning approaches quickly became the state-of-the-art in semantic segmentation.
Significant improvements were made by Long et al. \cite{Long2015}, who proposed a fully convolutional network  for pixel-wise semantic segmentation which resembles an encoder / decoder structure with skip connections to combine strong semantic features with spatially dense features.
This idea of feature processing at different scales was later refined by Lin et al. \cite{Lin2017}.
Their Feature Pyramid Network (FPN) adds features via 1x1 convolutions from the top-down path to features from the bottom-up path and is a common structure for many semantic segmentation and object detection models.
Inspired by FPN, Chen et al. developed DeepLab (e.g. \cite{Chen2018}), a modular development suite.
It uses atrous convolutions, batch normalization and can use different feature extractors such as Xception \cite{Chollet2017} or MobileNet \cite{Howard2019}.
In addition to depth-wise separable convolutions, which are also used in Xception, recent MobileNet variants use blocks of narrow-wide-narrow convolutions - termed inverted residuals - with linear residuals and a general structure optimized for bounded memory and computation.

Before semantic range measurement labels were available, semantic segmentation in grid maps was performed by projecting camera image labels onto a ground plane, e.g. via inverse perspective mapping \cite{Erkent2018, Lu2019}.
These label transfer methods however do not perform well due to imperfect sensor calibration, parallax and a violated plane assumption for larger distances.
Due to the sparse mapping of range measurement end points, some grid map layers are sparsely occupied.
Jaritz et al. \cite{Jaritz2018} propose ways to resolve issues with sparse observations such as removing batch norm in the first layer and using a validity mask.
\section{DATA SET GENERATION}
\label{sec:DATASETGENERATION}

\begin{figure}[b]
    \centering
    \includegraphics[width =\linewidth]{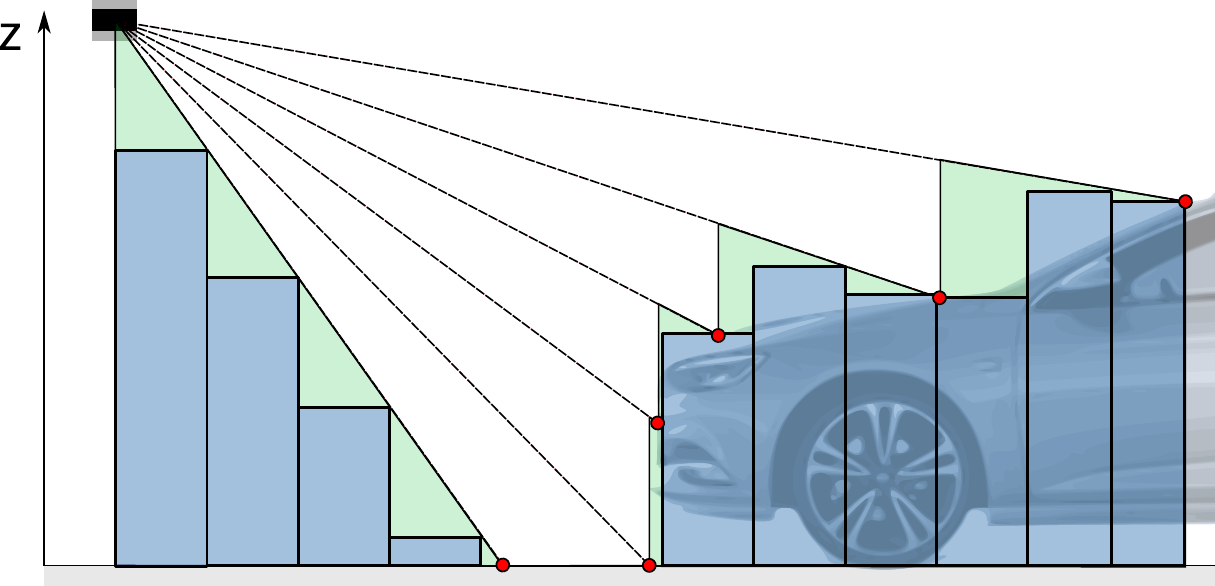}
    \caption{
        Longitudinal ray geometry of a multi-layer LiDAR sensor.
        Ray casting from the sensor origin to point detections (red) determines the potentially occupied space above ground (green), which is registered as a single z-value in layer \textit{Min. observed height} (blue).
    } 
    \label{fig:z_coord}
    \end{figure}

\begin{figure*}[thbp]
\centering
\includegraphics[width =0.95\textwidth]{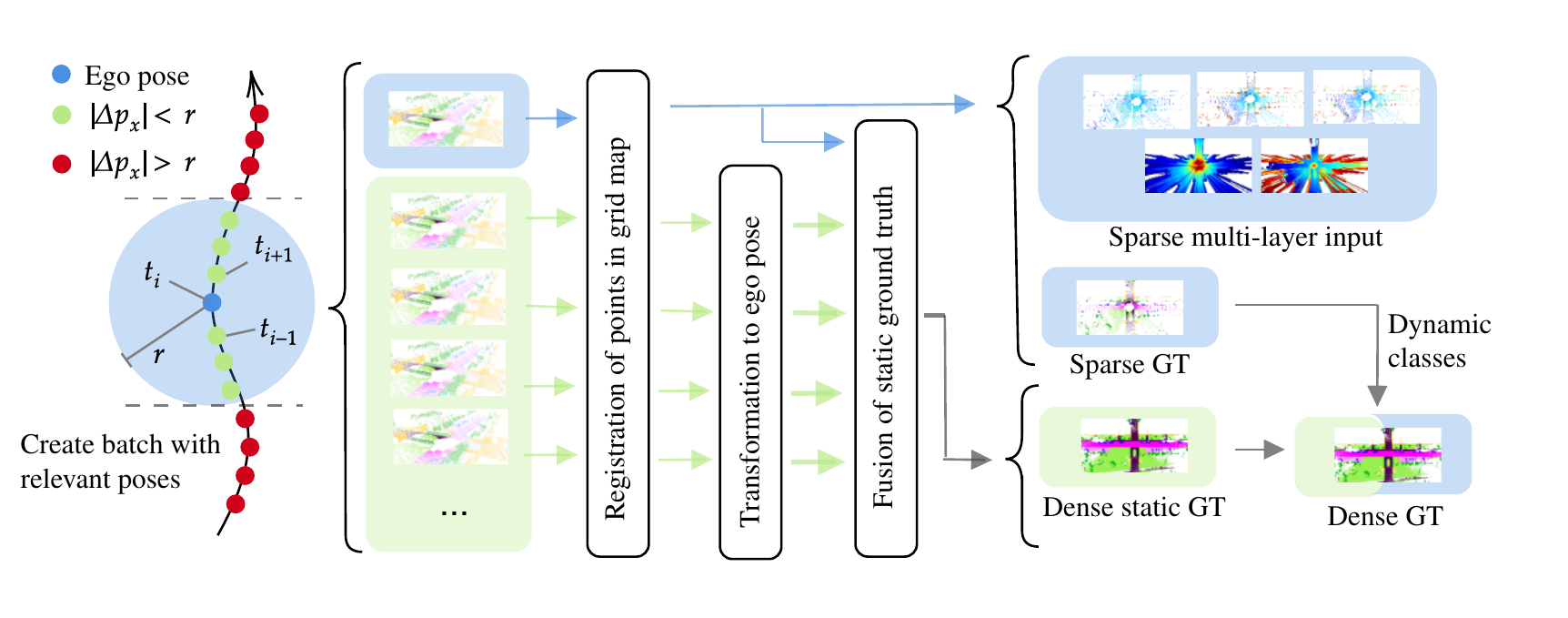}
\caption{
Each pose (blue) is separately processed by the framework resulting in five single-shot input layer and one sparse ground truth (sparse GT). Additionally, the framework determine the surrounding poses (green) within a certain radius ($r$). By assuming known poses, the framework fuses the static semantic information of all surrounding poses into one common grid map. This dense, static GT is finally enriched by the \textit{moving} labels of the sparse GT. 
}
\label{fig:poles}
\end{figure*}

\subsection{Grid Map Framework}
Each semantically labeled point cloud is mapped into a multi-layer grid map containing the layers depicted in Fig.~\ref{fig:IntroGridMapLayer}.
The following sparse layers are used:
\begin{itemize}
    \item \textbf{Intensity}, containing the average overall reflection intensities in a given grid cell.
    \item \textbf{Min. detected height}, height of the lowest point detection.
    \item \textbf{Max. detected height}, height of the highest point detection.
\end{itemize}

In addition to the sparse layers, which can be directly deduced from the point detections, rays are cast between the sensor origin and the point detections to obtain the following dense layers:
\begin{itemize}
    \item \textbf{Observability}, number of transmissions in a cell.
    \item \textbf{Min. observed height}, height of the lowest laser beam traversing the cell, see Fig.~\ref{fig:z_coord}.
\end{itemize}

In order to facilitate parallel computation and account for geometric sensor characteristics, all layers are first computed in polar coordinates and subsequently remapped into a cartesian coordinate system utilizing the texture mapping described in \cite{Homm2010}.

\subsubsection{Grid Resolution and Perceiving Distance}
A 2:1 grid map aspect ratio is used, oriented in such a way that the ego vehicle appears to be driving to the right.
This aspect ratio and orientation corresponds more accurately with the spatial distribution of detections in urban scenarios than with a quadratic grid map.
Each grid cell represents a region of 10cm$\times10$cm and each image has a resolution of $1001\times501$ pixels.
This corresponds to a region of interest of 100m$\times50$m with the sensor located in its center.

\subsubsection{Split of Data Set}
As suggested in the SemanticKITTI paper, we select the sequences 0-7 and 9-10 with a total number of 19130 scans for the training set and use the sequence 8 for model evaluation.
In contrast to the authors in \cite{Behley2019} we evaluate on the full evaluation set of 4071 scans to get a meaningful evaluation.
We argue, that we require a larger set, as our method reduces the amount of labeled points.

\subsubsection{Label Set Definition}

In \cite{Behley2019}, it is suggested to use 19 of 28 classes are for training. In the course of this work, the label set comprising 19 SemanticKITTI classes is further reduced in order to counter false classification of classes with similar appearance in a grid map and under-representation of rare classes.
The classes \textit{car, truck} and \textit{other-vehicle} are labeled as \textit{vehicle}, the classes \textit{traffic-sign}, \textit{fence} and \textit{pole} as \textit{object}, the classes \textit{bicyclist} and \textit{motorcyclist} as \textit{rider}, the classes \textit{bicycle} and \textit{motorcycle} as \textit{two-wheel} and the classes \textit{other-ground} and \textit{parking} as \textit{other-ground}.
This leaves 12 relevant classes and an additional class called \textit{unlabeled}.
All pixels labeled with the latter are excluded from the training process by ignoring them in the loss function. 

\subsubsection{Encoding of Ground Truth}
Next, we consider the label histogram for each grid cell individually and determine the cell class $k_\textit{i}$ by calculating a weighted arg max:
\begin{align}
k_{\mathrm{i}} = \underset{k \in K}{\mathrm{argmax}} \: w_\mathrm{k} \:  n_{\mathrm{i,k}},
\end{align}
where $n_{\mathrm{i,k}}$ is the total number of points of class $\mathrm{k}$  assigned to grid cell $\mathrm{i}$ and $w_\mathrm{k}$ is the class-specific weighting factor.

We choose the class-specific weighting factors as
\begin{align}
\label{eq:matching_costs}
w_\mathrm{k} = 
\begin{cases}
0 & \text{for \textit{unlabeled} }\\
1 & \text{for \textit{others} }\\
5 & \text{for \textit{vehicle}, \textit{person}, \textit{rider}, \textit{two-wheel}}\\
\end{cases}
\text{.}
\end{align}
The weighting factor for traffic participants is higher than for other classes.
This is a compromise between favoring traffic participants over other classes, if there is evidence for their presence and preventing distortion of their appearance.
Since the weighting factor of \textit{unlabeled} is zero, the label is never chosen for a grid cell if any point is assigned to it.
If no point is assigned to a grid cell, it is labeled as \textit{unlabeled}. 

The resulting label distribution is shown in Table \ref{tab::statistics}. Even after class-dependent weighting, the data set shows a significant under-representation of vulnerable road users in the data set.

\begin{table}[!htbp]
    \centering
        \caption{    
    Class distribution of semantic grid maps. A comparison of the label distributions for the training and evaluation set for both the sparse and dense ground truth.}
    \label{tab::statistics}
    \begin{tabular}{lcccc}
        \hline
	{}   &  \multicolumn{4}{c}{label distribution}\\
	{}   &  \multicolumn{2}{c}{sparse}&  \multicolumn{2}{c}{dense}\\     	
     	\cmidrule(lr){2-3} \cmidrule(lr){4-5}
        label & {$\DS{train}$} & {$\DS{val}$} & {$\DS{train}$}& {$\DS{val}$}\\ 
        \hline
vehicle		&	0.275	\% & 	0.353	\% & 	0.844	\% & 	1.070	\%\\
person		&	0.004	\% & 	0.011	\% & 	0.019	\% & 	0.047	\%\\
two-wheel 	&	0.002	\% & 	0.006	\% & 	0.007	\% & 	0.011	\%\\
rider		&	0.001	\% & 	0.004	\% & 	0.001	\% & 	0.006	\%\\
road		&	2.347	\% & 	2.029	\% & 	14.842	\% & 	12.942	\%\\
sidewalk	&	1.549	\% & 	1.277	\% & 	9.883	\% & 	8.444	\%\\
other-ground&	0.222	\% & 	0.153	\% & 	1.421	\% & 	1.022	\%\\
building	&	0.472	\% & 	0.471	\% & 	1.739	\% & 	1.988	\%\\
object		&	0.395	\% & 	0.110	\% & 	1.101	\% & 	0.399	\%\\
vegetation	&	2.043	\% & 	2.016	\% & 	6.284	\% & 	6.205	\%\\
trunk		&	0.025	\% & 	0.046	\% & 	0.062	\% & 	0.122	\%\\
terrain		&	1.374	\% & 	2.312	\% & 	7.717	\% & 	16.264	\%\\
unlabeled	&	91.290	\% & 	91.211	\% & 	56.078	\% & 	51.480	\%\\
        \hline
    \end{tabular}
\end{table}

\subsection{Dense Ground Truth}
The highly accurate sensor poses from the SemanticKITTI data set are used to enhance the ground truth density for each time point.
First, we create a batch of point clouds for each scan of a sequence which contains all point clouds of this sequence that possibly contribute to the semantic ground truth corresponding to the given scan.
For this, all point clouds with a sensor distance $|\Delta p_x|$ smaller than twice the maximum scan range $r$ are considered.
This leads to 40 to 300 aggregated point clouds per scan depending on the current velocity of the vehicle.
Highly accurate SLAM poses are required, as the GNSS poses from KITTI lead to alignment errors in the aggregated points cloud which make the semantic ground truth inconsistent.

Since moving objects cannot be aggregated over time using this method, only information from the current scan is used for this.
A sparse label map, consisting of points from all moving objects in the current scan, is superimposed on the dense grid map, keeping the knowledge of moving objects from the current scan.

\begin{table*}[!htbp]
\centering
 \caption{Quantitative results on the full evaluation set.}\label{tab::exp_results}
\label{my-label}
\begin{tabular}{llr|*{11}{p{0.75cm}}cc}
\toprule
&Approach &\textbf{mIoU}&vehicle&person&two-wheel&rider&road&side-walk&other\par-gro.&building&object&vege-tation&trunk&terrain\\
\midrule
\multirow{9}{*}{\rotatebox{90}{Sparse Eval}}
&$\mathbf{m3L_{i}}$& \textbf{0.325} & 0.455 & 0.0 & 0.0 & 0.0 & 0.810 & 0.532 & 0.114 & 0.649 & 0.117 & 0.593 & 0.000 & 0.626 \\
&$\mathbf{x41_{i}}$& \textbf{0.328} & 0.444 & 0.0 & 0.0 & 0.0 & 0.812 & 0.549 & 0.123 & 0.655 & 0.134 & 0.605 & 0.000 & 0.610  \\
&$\mathbf{x65_{i}}$& \textbf{0.324} &0.431 & 0.0 & 0.0 & 0.0 & 0.818 & 0.556 & 0.163 & 0.657 & 0.108 & 0.593 & 0.000 & 0.560 \\
&$\mathbf{m3L_{id}}$&\textbf{ 0.364} & 0.634 & 0.0 & 0.0 & 0.0 & 0.835 & 0.556 & 0.141 & 0.662 & 0.138 & 0.670 & 0.043 & 0.685   \\
&$\mathbf{x41_{id}}$& \textbf{0.363} & 0.650 & 0.0 & 0.0 & 0.0 & 0.837 & 0.569 & 0.165 & 0.687 & 0.146 & 0.653 & 0.009 & 0.642 \\
&$\mathbf{x65_{id}}$& \textbf{0.373} & 0.675 & 0.0 & 0.0 & 0.0 & 0.846 & 0.584 & 0.181 & 0.703 & 0.156 & 0.651 & 0.041 & 0.637  \\
&$\mathbf{m3L_{ido}}$& \textbf{0.373} & 0.661 & 0.0 & 0.0 & 0.0 & 0.836 & 0.554 & 0.150 & 0.703 & 0.142 & 0.665 & 0.100 & 0.666  \\
&$\mathbf{x41_{ido}}$& \textbf{0.382} & 0.689 & 0.0 & 0.0 & 0.0 & 0.852 & 0.587 & 0.209 & 0.708 & 0.154 & 0.664 & 0.065 & 0.653   \\
&$\mathbf{x65_{ido}}$& \textbf{0.398} & 0.697 & 0.0 & 0.0 & 0.0 & 0.858 & 0.603 & 0.259 & 0.728 & 0.151 & 0.689 & 0.099 & 0.693  \\
\addlinespace
\multirow{9}{*}{\rotatebox{90}{Dense Eval}}
&$\mathbf{m3L_{i}}$& \textbf{0.243} & 0.213 & 0.0 & 0.0 & 0.0 & 0.777 & 0.424 & 0.079 & 0.398 & 0.066 & 0.390 & 0.000 & 0.569  \\
&$\mathbf{x41_{i}}$&\textbf{ 0.260 }& 0.253 & 0.0 & 0.0 & 0.0 & 0.794 & 0.456 & 0.095 & 0.456 & 0.084 & 0.395 & 0.000 & 0.591   \\
&$\mathbf{x65_{i}}$& \textbf{0.246 }& 0.238 & 0.0 & 0.0 & 0.0 & 0.802 & 0.454 & 0.091 & 0.424 & 0.058 & 0.360 & 0.000 & 0.525  \\
&$\mathbf{m3L_{id}}$& \textbf{0.291} &0.408 & 0.0 & 0.0 & 0.0 & 0.807 & 0.460 & 0.128 & 0.450 & 0.100 & 0.488 & 0.035 & 0.620   \\
&$\mathbf{x41_{id}}$& \textbf{0.298} & 0.408 & 0.0 & 0.0 & 0.0 & 0.818 & 0.481 & 0.133 & 0.515 & 0.106 & 0.472 & 0.005 & 0.634 \\
&$\mathbf{x65_{id}}$& \textbf{0.303} & 0.430 & 0.0 & 0.0 & 0.0 & 0.826 & 0.498 & 0.155 & 0.521 & 0.102 & 0.459 & 0.028 & 0.622  \\
&$\mathbf{m3L_{ido}}$& \textbf{0.303} & 0.430 & 0.0 & 0.0 & 0.0 & 0.819 & 0.467 & 0.124 & 0.511 & 0.101 & 0.477 & 0.081 & 0.630\\
&$\mathbf{x41_{ido}}$& \textbf{0.312} & 0.435 & 0.0 & 0.0 & 0.0 & 0.835 & 0.494 & 0.157 & 0.547 & 0.112 & 0.486 & 0.040 & 0.638 \\
&$\mathbf{x65_{ido}}$& \textbf{0.328} & 0.433 & 0.0 & 0.0 & 0.0 & 0.843 & 0.514 & 0.229 & 0.547 & 0.108 & 0.510 & 0.063 & 0.686 \\

\bottomrule
\end{tabular}
\end{table*}

\section{EXPERIMENTS}
\label{sec:EXPERIMENTS}

In this work, we present the results of nine different deep learning models, which are all based on the DeeplabV3 \cite{Chen2018} architecture. This section describes the general experimental setup and how the nine models differ.

\subsection{Experimental Setup}

\subsubsection{General}
In the training process, we purely consider sparse, single-shot laser scans - this holds for both input data and for the ground truth. For all experiments, we train on the full grid map resolution of $1001 \times 501$ pixels and use the full label map of 12 classes as shown in Table~\ref{tab::statistics}. Due to GPU limitations, we choose the batch size of the Xception models to two and trained the MobileNet model with a batch size of four. We trained all models for at least 200000 iterations. In the training process, all unlabeled ground truth cells are excluded from the optimization. The loss is scaled with respect to the share of unlabeled cells. %

\subsubsection{Data Augmentation}
In deep learning applications, augmentation methods avoid the network to overfit on the training data. Hence, they are considered to be crucial for the learning process as they increase the model's performance on the evaluation or test set \cite{Shorten2019}. In this work, we apply multiple augmentation methods to prevent the network from overfitting. The training data was randomly flipped  along the vertical axis and randomly scaled with a factor between $0.8$ and $1.2$. %
 \cite{Srivastava2014}.
 
\subsubsection{Initialization}
Generally, literature recommends to initialize a deep convolutional neural network instead of training it from scratch \cite{IanGoodfellowYoshuaBengio2015}. In this work, we use weights, pre-trained on ImageNet \cite{Krizhevsky2017}, which were released in the official DeepLab repository \cite{Chen2018}. As our input has up to five channels, we randomly initialize the weights of the very first layer. This is done for every number of input channels to ensure comparability between our experiments.

\subsection{Model Outline}
The models differ in the used backbone architecture and the amount of input channels which are also influencing the first network layers.

\subsubsection{Feature Extractor}
We consider three different feature extractors as backbone for the DeepLabV3 architecture. Two of them are based on the Xception model, a powerful, yet relative efficient network. It uses depthwise separable convolutions and is based on the empirically proven hypothesis of the inception modules, i.e. mapping the cross-channel correlation separately from the spatial correlations is effective and efficient. We refer to them as $\mathbf{x41}$ for the model with 41 layers and $\mathbf{x65}$ for the larger 65-layered network. Further, we chose the MobileNetV3-large \cite{Howard2019}, later called $\mathbf{m3L}$ because it is considered a state-of-the-art-network in terms of efficiency. %

\subsubsection{Definiton of Input Layer}
To fully evaluate the benefit of a multi-layer grid map representation, we choose three different input configurations, which are denoted by an index as follows:
\begin{itemize}
\item $\mathbf{i}$: Using \textit{intensity} as single layer input.
\item $\mathbf{id}$: Adding the height information, \textit{Min. detected height} and \textit{Max. detected height} to the  \textit{intensity} layer, resulting in three input channels.
\item $\mathbf{ido}$: Expanding the input size to five by adding the dense layers \textit{Observability} and \textit{Min. observed height} to the three layers above. 
\end{itemize}

\section{EVALUATION}
\label{sec:EVALUATION}

\subsection{Metrics}
We choose the \textit{Intersection over Union} (IoU) \cite{Everingham2014}, also known as \textit{Jaccard index}, for the qualitative evaluation as it is a commonly used measure for semantic segmentation. The per-class IoU$_k$ is calculated by
\begin{equation}
\text{IoU}_k = \frac{T_{P_k} }{T_{P_k}  + F_{P_k}  +F_{N_k} },
\end{equation}
with k being one of 12 classes. Consequently, the  \textit{mean Intersection over Union}, mIoU, is calculated as follows:
\begin{equation}
\text{mIoU} = \frac{1}{|K|} \sum_{k \in K} \text{IoU}_k
\end{equation}
where $| K|$ is the the labelset's cardinality.

\subsection{Evaluation Modes}
Each experiment is evaluated using two different approaches. Note that the network always predicts a dense semantic grid map, independent of the evaluation method. For the first one, we take the sparse ground truth, which is generated by processing a single scan. This ground truth corresponds to the ground truths, which are used for the training process. We denote this mode as \textit{Sparse Eval} evaluation. The second method, referred to \textit{Dense Eval} evaluation, considers the dense ground truth. Instead of evaluating with the full dense grid map, we filter it first with the observability input layer, so that occluded areas are removed from the evaluation. A filtered prediction map can be seen in the middle of Fig.~\ref{fig:qualitative_results}.

\subsection{Experimental Results} \label{sec::EVAL::ExperimentalResults}
\begin{figure*}[!htbp]
    \renewcommand{\thesubfigure}{\arabic{subfigure}}
    \centering
    \begin{subfigure}{\textwidth}
    \parbox[c]{.03\linewidth}{\subcaption{}}
    \parbox[c]{\linewidth}{
        \includegraphics[width = 0.95\textwidth ]{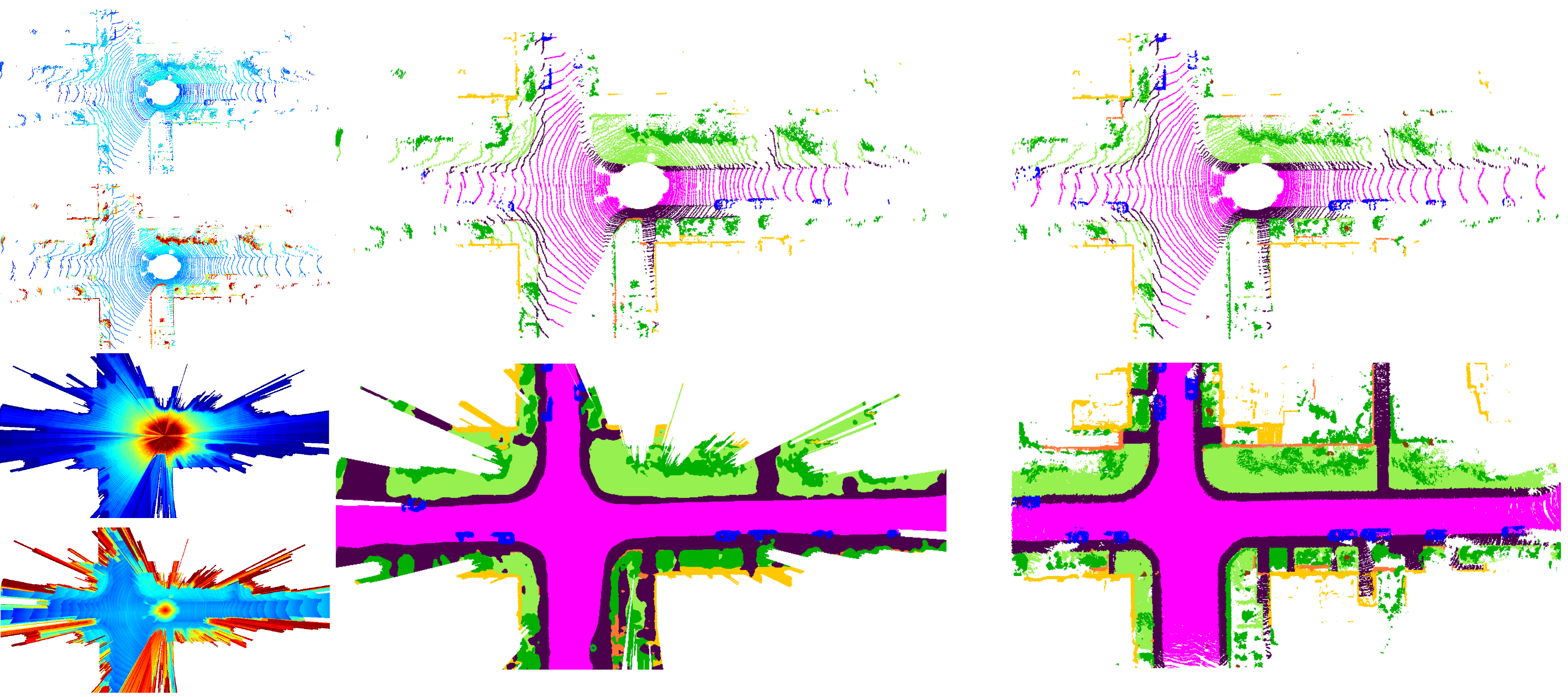}
    }
    \end{subfigure}
    \\ \vspace*{1.0cm}
    \begin{subfigure}{\textwidth}
        \parbox[c]{.03\linewidth}{\subcaption{}}
        \parbox[c]{\linewidth}{
            \includegraphics[width = 0.95\textwidth ]{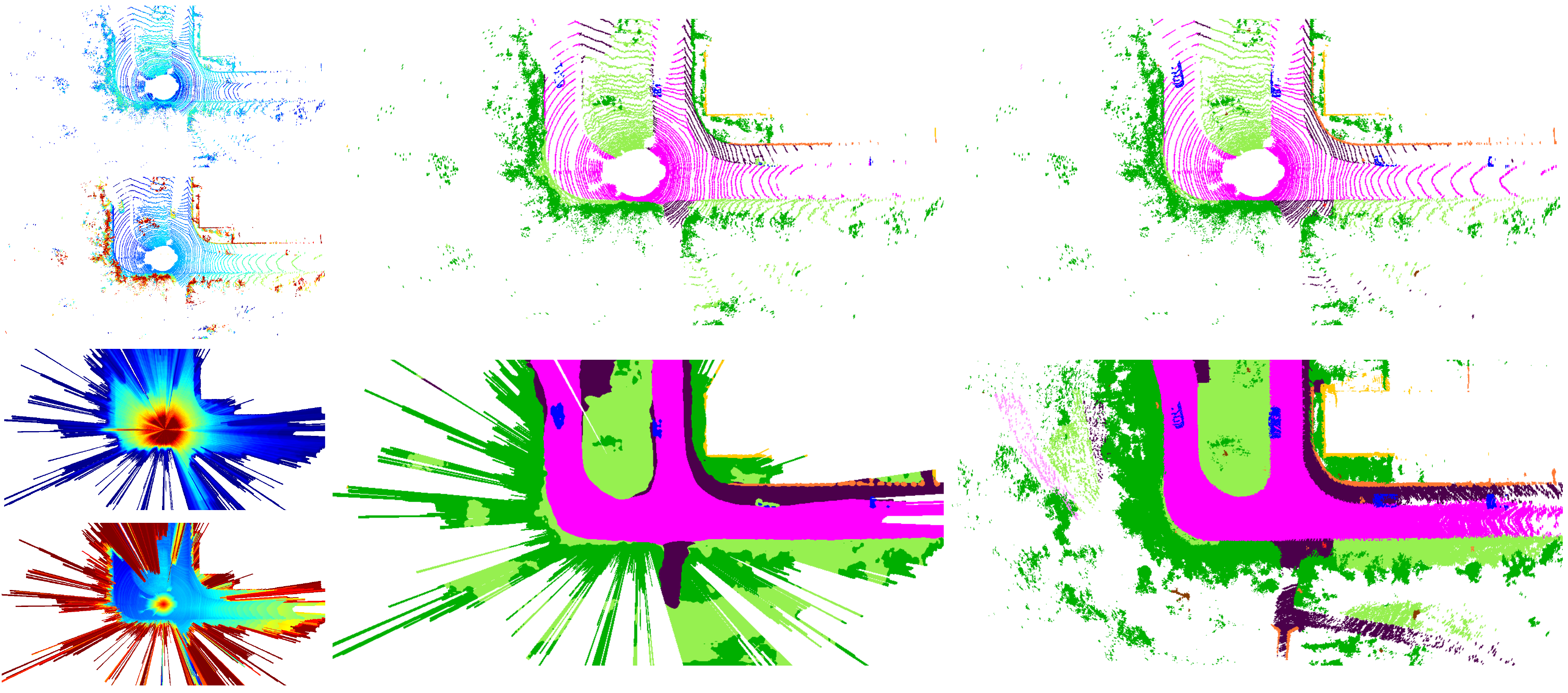}
        }
        \end{subfigure}
            \\
    \centering
    \caption{
        Qualitative results from the evaluation set, processed by $\mathbf{x65_{ido}}$\label{fig:augm_l1}.
        The network was trained on sparse, multi-layer input and sparse ground truth.
        We depict four out of five input layers in the left column - from top to down: \textit{intensity}, \textit{max. detected height}, \textit{observability} and \textit{min. observed height}. The middle column depicts the predictions as a sparse layer on top and a dense layer below, filtered by the observed area.
        The corresponding ground truth is depicted in the right column.
    }
    \label{fig:qualitative_results}
\end{figure*}
Table \ref{tab::exp_results} presents the mIoU scores obtained through the experiments. We observe high accuracy in the predictions of \textit{road}, \textit{sidewalk}, \textit{vegetation}, \textit{terrain} and \textit{building} up to an IoU of 85,8 \%. All of these classes are also well represented in the data set distribution (see Table \ref{tab::statistics}). In contrast, the three classes, which represent vulnerable road users cannot be learned at all. All of them are highly underrepresented, e.g. there are 2300 times more labels of the \textit{road} class than there are \textit{rider} labels. Another explanation might be that the grid map representation is not suitable for these classes. Furthermore, it can be observed that the Xception models outperform the MobileNet by a slight margin.

A closer look at the different input configurations reveals that a multi-layer input is highly beneficial for the task of semantic segmentation of grid maps. In the conducted experiments, both the additional height layers and the dense observation layers boost the IoU for each class. While some classes yield reasonable results with only the \textit{intensity} layer (e.g. \textit{road}, \textit{sidewalk}), the multi-layer approach seems to play a key role for others: For instance, the \textit{vehicle} class gains 20\% IoU in all experiments after adding the two height layers and \textit{trunk} seems to be unpredictable without the height information.

We now consider the results obtained in \textit{Dense Eval}. The IoUs do only slightly drop in comparison to \textit{Sparse Eval}, while at the same time a multitude of labels are available and considered in the evaluation. This means that many cells without corresponding detections in the input layers are still predicted correctly according to the dense ground truth. This evaluation shows that the grid map representation enables the dense prediction of semantic information, even if it was trained on sparse ground truth data.
 
Qualitatively, Fig.~\ref{fig:qualitative_results} shows the network's ability to accurately predict the general configuration of a road scene, the road course, drivable space and large, well-discernible objects such as vehicles or walls. Small or vertically aligned objects such as poles, trunks and especially pedestrians are predicted less accurately. The dense prediction from sparse input represents the scenery quite well with mostly smooth und continuous area shapes, which are similar to the dense ground truth.

\subsection{Inference Time}
We evaluated the processing times on an NVIDIA GeForce RTX 2080 Ti GPU with 11 GB graphics memory. The presented models are trained using the Tensorflow framework \cite{GoogleResearch2015} and executed in Python. Table \ref{tab::inference} summarizes the results. %

\begin{table}[h]
\centering
\caption {Comparison of model architectures with respect to inference time.}\label{tab::inference}
    \begin{tabular}{c@{\hspace{0.3cm}}ccc}
\toprule
     {}   &  \multicolumn{3}{c}{inference time in ms}\\
     \cmidrule(lr){2-4}
    {Input layers } & { $\mathbf{x41}$} & { $\mathbf{x65}$} & { $\mathbf{m3L}$}  \\ \midrule
    {$\mathbf{i}$} & $68,6$& $84,9$ &$32,7$\\
    {$\mathbf{id}$} & $69,2$& $85,7$ &$33.1$\\
    {$\mathbf{ido}$} & $70,0$& $87,3$ &$34,4$\\
    \bottomrule
\end{tabular}%
\end{table}

As shown in Section \ref{sec::EVAL::ExperimentalResults}, $\mathbf{m3L}$ yields classification results which are similar to the Xception networks. However, it is significantly faster than the $\mathbf{x41}$ and $\mathbf{x65}$ network, which makes it a promising network for mobile perception systems such as intelligent vehicles regarding our proposed classification task. We also see that the addition of more input layers has very little effect on the inference time, which can be explained by the fact that only the computational effort in the very first layer is influenced by the input's channel size.  %

\section{CONCLUSION}
\label{sec:CONCLUSION}
In this work, we presented an approach towards semantic grid map estimation based on sparse input data. Further we developed and presented a method to generate a dense semantic ground truth which can be used to refine and evaluate the network's output. By evaluating on sparse and dense ground truth, we demonstrated that our approach enables the learned prediction of dense labels given a sparse input. The experiments have shown that a multi-layer input boosts the performance regarding both sparse and dense prediction. While the algorithm achieves good classification results for most classes, it fails to predict vulnerable road users. This might be due to their under-representation in the created grid map data set.

As a next step, we will integrate the dense ground truth layer within the learning process to teach the model directly the mapping from sparse input to a dense representation. Furthermore, we aim to fuse sequential input data within the grid map framework to enrich the sparse input and enhance the semantic segmentation performance.

\section{Acknowledgment}
The  authors  thank  Daimler  AG  for  the  fruitful  collaboration and the support for this work. \printbibliography

\end{document}